\documentclass[11pt,a4paper]{article}
\usepackage[hyperref]{emnlp2020}
\usepackage{times}
\usepackage{latexsym}

\usepackage{multirow}
\usepackage{array} 
\usepackage{diagbox}
\usepackage{amssymb}
\usepackage{graphicx}

\graphicspath{ {./images/} }

\usepackage{microtype}
\usepackage{enumitem}
\usepackage{pifont}
\aclfinalcopy % Uncomment this line for the final submission
\def\aclpaperid{165} %  Enter the acl Paper ID here

\newcolumntype{P}[1]{>{\centering\arraybackslash}p{#1}} % px

\newcommand{\cmark}{\ding{51}}%
\newcommand{\xmark}{\ding{55}}%

\title{Towards Persona-Based Empathetic Conversational Models}

\author{Peixiang Zhong$^{1,2,3}$, Chen Zhang$^4$, Hao Wang$^4$, Yong Liu$^{1,3}$, Chunyan Miao$^{1,2,3}$\thanks{\hspace{0.18cm}Corresponding author} \\ 
$^1$Alibaba-NTU Singapore Joint Research Institute\\
$^2$School of Computer Science and Engineering\\
$^3$Joint NTU-UBC Research Centre of Excellence in Active Living for the Elderly\\
Nanyang Technological University, Singapore\\
$^4$Alibaba Group, China\\
\normalsize{\texttt{peixiang001@e.ntu.edu.sg}},  \normalsize{\texttt{zhangchen010295@163.com}},
\normalsize{\texttt{cashenry@126.com}},\\
\normalsize{\texttt{stephenliu@ntu.edu.sg}},
\normalsize{\texttt{ascymiao@ntu.edu.sg}}
}

\date{}

\begin{document}
\maketitle
\begin{abstract}
Empathetic conversational models have been shown to improve user satisfaction and task outcomes in numerous domains. In Psychology, persona has been shown to be highly correlated to personality, which in turn influences empathy. In addition, our empirical analysis also suggests that persona plays an important role in empathetic conversations. To this end, we propose a new task towards persona-based empathetic conversations and present the first empirical study on the impact of persona on empathetic responding. Specifically, we first present a novel large-scale multi-domain dataset for persona-based empathetic conversations. We then propose CoBERT, an efficient BERT-based response selection model that obtains the state-of-the-art performance on our dataset. Finally, we conduct extensive experiments to investigate the impact of persona on empathetic responding. Notably, our results show that persona improves empathetic responding more when CoBERT is trained on empathetic conversations than non-empathetic ones, establishing an empirical link between persona and empathy in human conversations.
\end{abstract}

\section{Introduction}
\label{sec: introduction}
% 1. what is empathy
% 2. why empathetic conversations
% 3. existing works on empathetic conversations and their limitations
% 4. why we incorporate persona into empathetic conversations
% 5. briefly introduce our research objective
% 6. briefly introduce our approach
% 7. briefly introduce our conclusion
% 8. briefly summarize our contributions
Empathy, specifically affective empathy, refers to the capacity to respond with an appropriate emotion to another's mental states \cite{rogers2007cares}. 
In NLP, empathetic conversational models have been shown to improve user satisfaction and task outcomes in numerous domains \cite{klein1998computer, liu2005embedded, wright2008empathy, fitzpatrick2017delivering, zhou2018design}. For example, empathetic agents received more positive user ratings, including greater likeability and trustworthiness than controls \cite{brave2005computers}.

In recent years, neural network based conversational models \cite{vinyals2015neural, lowe2015ubuntu} are becoming dominant. \citet{zhou2018design} designed XiaoIce, a popular AI companion with an emotional connection to satisfy the human need for communication, affection, and social belonging. Recently, \citet{rashkin2019towards} presented a new dataset and benchmark towards empathetic conversations and found that both Transformer-based generative models \cite{vaswani2017attention} and BERT-based retrieval models \cite{devlin2019bert} relying on this dataset exhibit stronger empathy.
\begin{figure}
    \centering
    \includegraphics[width=\linewidth]{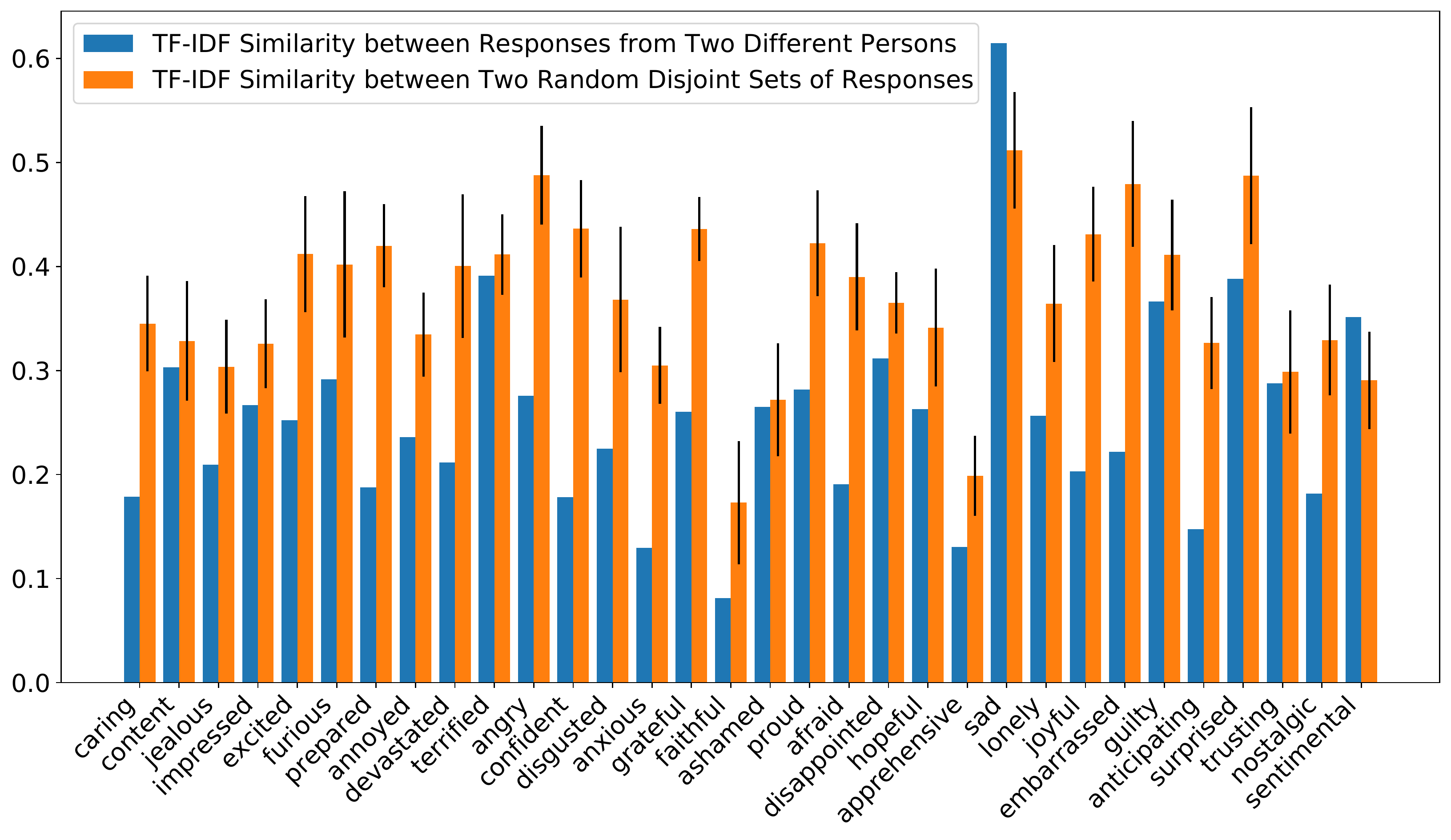}
    \caption{TF-IDF similarity between two sets of empathetic responses \cite{rashkin2019towards} for each emotion (best viewed in color). For most emotions (28 out of 32), the similarity between responses from two different speakers (blue) is substantially smaller than the similarity between two random disjoint sets of responses (orange, averaged over five runs).}
    \label{fig: tfidf}
\end{figure}

However, most existing studies, e.g., \cite{rashkin2019towards}, do not consider persona when producing empathetic responses\footnote{One exception is XiaoIce \cite{zhou2018design}, however, her persona is not configurable and thus difficult to satisfy various human needs.}.
In Psychology, persona refers to the social face an individual presents to the world \cite{jung2016psychological}. Persona has been shown to be highly correlated with personality \cite{leary2011personality}, which in turn influences empathy \cite{richendoller1994exploring, costa2014associations}. In addition, our empirical analysis of empathetic conversations in \cite{rashkin2019towards} also shows that for most emotions, the empathetic responses from two different persons\footnote{Each response in \cite{rashkin2019towards} has a speaker id but no persona.} have more differences than that between two disjoint sets of random responses, as shown in Figure \ref{fig: tfidf}. Both the theories in Psychology and the evidence from our empirical analysis suggest that persona plays an important role in empathetic conversations, which, to the best of our knowledge, has not been investigated before\footnote{A very recent work \cite{roller2020recipes} incorporates persona and empathy by fine-tuning on corresponding datasets, however, it does not investigate the impact of persona on empathetic responding.}. 

To this end, we propose a new task towards persona-based empathetic conversations and present the first empirical study on the impact of persona on empathetic responding. Our study would be beneficial to researchers in Dialogue Systems and Psycholinguistics. However, one major challenge of this study is the lack of relevant datasets, i.e., existing datasets only focus on either persona or empathy but not both (see Table \ref{table: dataset comparision} for details).
In this paper, we present a novel large-scale multi-turn \textbf{P}ersona-based \textbf{E}mpathetic \textbf{C}onversation (\textbf{PEC}) dataset in two domains with contrasting sentiments, obtained from the social media Reddit, to facilitate our study.

We then propose CoBERT, an efficient BERT-based response selection model using multi-hop co-attention to learn higher-level interactive matching. 
CoBERT outperforms several competitive baselines on PEC, including Poly-encoder \cite{humeau2020polyencoders}, the state-of-the-art BERT-based response selection model, by large margins. 
We conduct additional comparisons with several BERT-adapted models and extensive ablation studies to evaluate CoBERT more comprehensively. 

Finally, based on PEC and CoBERT, we investigate the impact of persona on empathetic responding. 
In addition, we analyze how limited persona data improves model performance, and how our model generalizes to new personas.

In summary, our contributions are as follows:
\begin{itemize}[leftmargin=*]
    \item We propose a new task and a novel large-scale multi-domain dataset, PEC, towards persona-based empathetic conversations. Our data and code are available here\footnote{https://github.com/zhongpeixiang/PEC}.
    \item We propose CoBERT, a BERT-based response selection model that obtains the state-of-the-art performance on PEC. Extensive experimental evaluations show that CoBERT is both effective and efficient.
    \item We present the first empirical study on the impact of persona on empathetic responding. The results show that persona improves empathetic responding \textit{more} when CoBERT is trained on empathetic conversations than non-empathetic ones, establishing an empirical link between persona and empathy in human conversations.
\end{itemize}

\section{Related Work}
\label{sec: related work}
\textbf{Empathetic Conversational Models} Despite the growing number of studies in neural conversational models, less attention has been paid to make conversations empathetic until recently \cite{siddique2017zara, morris2018towards, shi2018sentiment, lin2019caire, shin2019happybot, rashkin2019towards, li2019empgan, lin2019moel, zandie2020emptransfo}, possibly due to the lack of empathetic conversation datasets. \citet{rashkin2019towards} proposed EMPATHETICDIALOGUES (\textbf{ED}), the first empathetic conversation dataset comprising 25K conversations in 32 emotions. 
Conversational models trained on the role of the listener in the dataset exhibited stronger empathy than models trained on non-empathetic datasets. We compare ED and PEC in the last paragraph of Section \ref{sec: PEC dataset}.

\noindent\textbf{Persona-Based Conversational Models} In recent years, personalized conversational models are emerging \cite{li2016persona, zhang2018personalizing, wolf2019transfertransfo, chan2019modeling, madotto2019personalizing, zheng2019pre}. \citet{li2016persona} proposed persona embeddings in a response generation model and achieved improved generation quality and persona consistency. \citet{zhang2018personalizing} proposed PERSONA-CHAT (\textbf{PC}), a crowd-sourced conversation dataset with persona information, to improve model engagingness and consistency. \citet{mazare2018training} further presented a much larger persona-based conversation dataset collected from Reddit (\textbf{PCR}) and showed that persona consistently improves model performance even when a large number of conversations is available for training. We compare PC, PCR, and PEC in the last paragraph of Section \ref{sec: PEC dataset}.
Recently, \citet{gu2019dually} proposed DIM, a personalized response selection model with interactive matching and hierarchical aggregation, and achieved state-of-the-art performance on PC.

\noindent\textbf{Retrieval-based Conversational Models} Recent neural retrieval-based conversational models generally have three modules: encoding, matching and aggregation \cite{lowe2015ubuntu, zhou2016multi, wu2017sequential, zhou2018multi, zhang2018modeling, chen2019sequential, feng2019learning, yuan2019multi}. The encoding module encodes text into vector representations using encoders such as LSTM, Transformer, or BERT. The matching module measures context-response associations using various attention mechanisms at different granularities. The aggregation module summarizes the matching information along the sequence dimension to obtain the final representation. A recent work \citet{humeau2020polyencoders} proposed Poly-encoder, an efficient BERT-based response selection model that obtained the state-of-the-art performance on multiple conversation datasets.

\section{The PEC Dataset}
\label{sec: PEC dataset}
In this section, we introduce the collection procedure and statistics of our proposed persona-based empathetic conversation (PEC) dataset.
\begin{table}[!t]
\small
\centering
\begin{tabular}{p{1.2cm}p{.55cm}p{.55cm}p{.55cm}p{.55cm}p{.55cm}p{.55cm}}
\hline
 & \multicolumn{3}{c}{\textbf{happy}} & \multicolumn{3}{c}{\textbf{offmychest}}\\
\hline
 & train & valid & test & train & valid & test\\
\hline
\#Conv. & 157K & 20K & 23K & 124K & 16K & 15K\\
% \hline
\#Utter. & 367K & 46K & 54K & 293K & 38K & 35K\\
% \hline
\#Speaker & 93K & 17K & 19K & 89K & 16K & 16K\\
\hline
\#Avg.PS & 66.0 & 70.8 & 70.0 & 59.6 & 66.8 & 67.1\\
% \hline
\#Std.PS & 38.1 & 36.7 & 36.9 & 40.2 & 39.0 & 38.8\\
\hline
\#Avg.U & 21.5 & 21.9 & 21.3 & 30.4 & 31.5 & 30.0\\
% \hline
\#Avg.P & 10.9 & 10.8 & 10.8 & 10.9 & 10.9 & 10.9\\
\hline
\end{tabular}
\caption{Statistics of PEC. \#Avg.PS and \#Std.PS denote average and standard deviation of the number of persona sentences per speaker, respectively. \#Avg.U denotes the average utterance length. \#Avg.P denotes the average persona sentence length.}
\label{table: datasets}
\end{table}

\begin{table}[!t]
\small
\centering
\begin{tabular}{cccc}
\hline
 & \textbf{happy} & \textbf{offmychest} & \textbf{control group}\\
\hline
Sentiment & 0.85 & -0.39 & 0.03\\
\hline
Empathy & 0.73 & 0.61 & 0.25\\
\hline
\end{tabular}
\caption{Sentiment and empathy of PEC and the control group based on human ratings. Sentiment ranges from -1 (negative) to 1 (positive). Empathy ranges from 0 (non-empathetic) to 1 (empathetic). Ratings are aggregated by majority voting (averaging shows similar results). The inter-annotator agreement, measured by Fleiss' kappa \cite{fleiss1971measuring}, for sentiment and empathy are 0.725 and 0.617, respectively. Both agreement statistics indicate ``substantial agreement".}
\label{table: dataset annotations}
\end{table}

\noindent\textbf{Data Source} We collect empathetic conversations from two subreddits \textit{happy}\footnote{https://www.reddit.com/r/happy/} and \textit{offmychest}\footnote{https://www.reddit.com/r/offmychest/} on Reddit, a discussion forum where users can discuss any topics on their corresponding sub-forums/subreddits. The \textit{happy} subreddit is where users share and support warm and happy stories and thoughts. The \textit{offmychest} subreddit is where users share and support deeply emotional things that users cannot tell people they know. We choose these two subreddits as our data source because their posts have contrasting sentiments and their comments are significantly more empathetic than casual conversations, i.e., the control group, as shown in Table \ref{table: dataset annotations}. 

\noindent\textbf{Conversation Collection} Discussions on Reddit are organized in threads where each thread has one post and many direct and indirect comments. Each thread forms a tree where the post is the root node and all comment nodes reply to their parent comment nodes or directly to the root node. Therefore, given a thread with $n$ nodes, we can extract $n-1$ conversations where each conversation starts from the root node and ends at the $n-1$ non-root nodes. We randomly split conversations by threads according to the ratio of 8:1:1 for training, validation, and test sets, respectively. 
\begin{table*}[!t]
\small
\centering
\begin{tabular}{p{0.4cm}|p{5.2cm}|p{8.8cm}}
\hline
 & \textbf{happy} & \textbf{offmychest}\\
\hline
% \multirow{5}{*}{Conversation}
 & Celebrating 43 years of marriage with the love of my life.& Worried. Am I becoming depressed again? Please don't leave me. Is everything okay? You don't seem yourself.\\
\cline{2-3}
\parbox[t]{2mm}{\multirow{5}{*}{\rotatebox[origin=c]{90}{Conversation}}} & She looks very young for someone who has been married 43 years. That must surely put her in the 63-73yr age range?! & I'm living these exact words. \\
\cline{2-3}
 & I just turned 61, thanks! & I hope everything works out for you. I'm trying not to fall apart. \\
\cline{2-3}
 & I hope I look that young when I'm 61! You guys are too cute, congratulations :) & Me too. If you ever want someone to talk to my messages are open to you. \\
\hline
\parbox[t]{2mm}{\multirow{4}{*}{\rotatebox[origin=c]{90}{Persona}}}  & I took an 800 mg Ibuprofen and it hasn't done anything to ease the pain. &  I think I remember the last time I ever played barbies with my litter sister.\\
\cline{2-3}
 & I like actively healthy. & I have become so attached to my plants and I really don't want it to die. \\
\cline{2-3}
& I want a fruit punch! & I'm just obsessed with animals. \\
\hline
\end{tabular}
\caption{Two example conversations with personas from PEC. The persona sentences correspond to the last speakers in the conversations.}
\label{table: examples}
\end{table*}

\noindent\textbf{Persona Collection} Following \cite{mazare2018training}, for each user in the conversations, we collect persona sentences from all posts and comments the user wrote on Reddit. The posts and comments are split into sentences, and each sentence must satisfy the following rules to be selected as a persona sentence: 1) between 4 and 20 words; 2) the first word is ``i"; 3) at least one verb; 4) at least one noun or adjective; and 5) at least one content word. Our rules are stricter than that from \cite{mazare2018training}, allowing us to extract less noisy persona sentences. For each user, we extract up to 100 persona sentences. 

Note that we choose our approach to persona collection because 1) the well-established work \cite{mazare2018training} successfully trained personalized agents using this approach; 2) this approach is significantly more scalable and cost-effective than crowd-sourcing; and 3) we are concerned that using crowd-sourcing, i.e., assigning artificial personas to crowd-workers and asking them to chat empathetically based on the assigned personas, would introduce worker-related noises such that models may merely learn superficial empathetic responding patterns that crowd-workers deem suitable given the assigned personas.

\begin{table}[!t]
\small
\centering
\begin{tabular}{P{1.4cm}P{0.7cm}P{0.8cm}P{0.9cm}P{0.6cm}P{0.6cm}}
% \begin{tabular}{c|c|c|c|c|c}
\hline
Dataset & Source & Persona & Empathy & Size & Public\\
\hline
ED & CS & \xmark & \cmark & 78K & \cmark  \\
% \hline
PC & CS & \cmark & \xmark & 151K & \cmark  \\
% \hline
PCR & Reddit & \cmark & \xmark & 700M & \xmark  \\
\hline
PEC (ours) & Reddit & \cmark & \cmark & 355K & \cmark  \\
\hline
\end{tabular}
\caption{Comparisons between PEC and related datasets. ED denotes EMPATHETICDIALOGUES \cite{rashkin2019towards}. PC denotes PERSONA-CHAT \cite{zhang2018personalizing}. PCR denotes the persona-based conversations from Reddit \cite{mazare2018training}. CS denotes crowd-sourced. The size denotes the number of expanded conversations.}
\label{table: dataset comparision}
\end{table}

\noindent\textbf{Data Processing} We keep a maximum of 6 most recent turns for each conversation. We filter conversations to ensure that 1) each post is between 2 and 90 words; 2) each comment is between 2 and 30 words\footnote{Posts are usually longer than comments. 87\% posts and 82\% comments on \textit{happy} are less than 90 and 30 words, respectively. 24\% posts and 59\% comments on \textit{offmychest} are less than 90 and 30 words, respectively.}; 3) all speakers have at least one persona sentence; and 4) the last speaker is different from the first speaker in each conversation. The last requirement is to maximally ensure that the last utterance is the empathetic response instead of a reply of the poster. In addition, persona sentences appearing in the conversation responses are removed to avoid data leakage. Finally, we lower-case all data and remove special symbols, URLs, and image captions from each sentence. The statistics of PEC are presented in Table \ref{table: datasets}. Two examples of PEC are shown in Table~\ref{table: examples}.

Note that it may not be easy to see explicit links in Table 3, but that’s exactly what we are studying for, i.e., to uncover the implicit (and possibly unexpected) links between persona and empathy using real user data. For example, the utterance “I hope I look that young” may implicitly link to the persona “I like actively healthy” in Table \ref{table: examples}.

\noindent\textbf{Data Annotations} We manually annotate 100 randomly sampled conversations from each domain to estimate their sentiment and empathy. 
To avoid annotation bias, we add a control group comprising 100 randomly sampled casual conversations from the \textit{CasualConversation}\footnote{https://www.reddit.com/r/CasualConversation/} subreddit, where users can casually chat about any topics. Finally, we mix and shuffle these 300 conversations and present them to three annotators. The annotation results are presented in Table \ref{table: dataset annotations}. The posts in the happy and offmychest domains are mostly positive and negative, respectively. Both domains are significantly more empathetic than the control group ($p<0.001$, one-tailed $t$-test).

\noindent\textbf{Conversation Analysis}
We conduct conversation analysis for PEC, similar to our analysis for ED \cite{rashkin2019towards} in Figure \ref{fig: tfidf}. Specifically, the TF-IDF similarities between responses from two different persons are 0.25 and 0.17 for happy and offmychest, respectively, whereas the TF-IDF similarities between two disjoint sets of random responses are 0.38 ($\pm$0.05) and 0.31 ($\pm$0.05) for happy and offmychest over 5 runs, respectively. The results show that empathetic responses between different persons are more different than that between random empathetic responses in PEC, suggesting that different speakers in PEC have different ``styles" for empathetic responding.

\noindent\textbf{Comparisons with Related Datasets} Table \ref{table: dataset comparision} presents the comparisons between PEC and related datasets. PEC has the unique advantage of being both persona-based and empathetic. In addition, PEC is collected from social media, resulting in a much more diverse set of speakers and language patterns than ED \cite{rashkin2019towards} and PC \cite{zhang2018personalizing}, which are collected from only hundreds of crowd-sourced workers. Finally, PEC is over 2x larger than the other two public datasets, allowing the exploration of larger neural models in future research.

\section{Our CoBERT Model}
\label{sec: CoBERT}

\begin{figure}
    \centering
    \includegraphics[width=\linewidth]{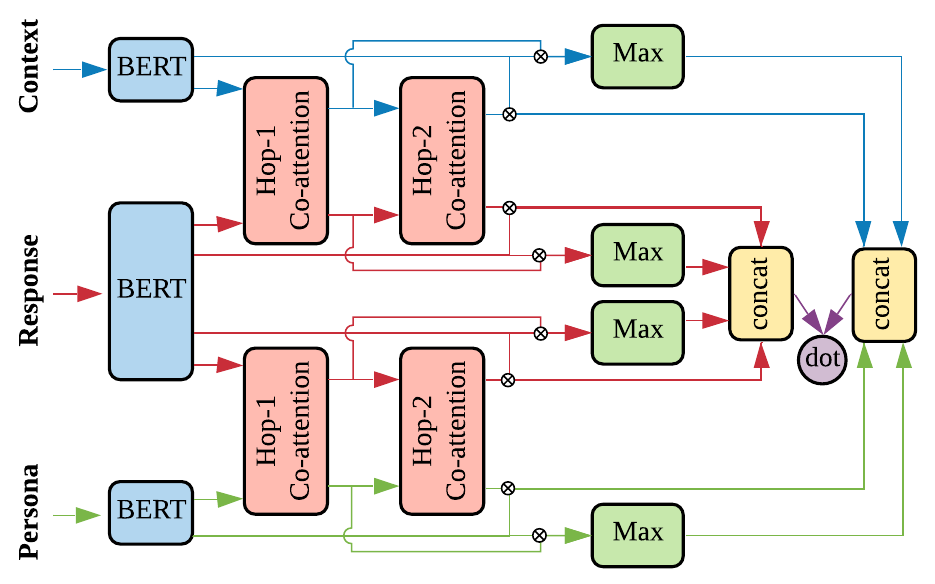}
    \caption{Our CoBERT architecture.}
    \label{fig: cobert}
\end{figure}

In this section, we briefly introduce the task of response selection and present our proposed CoBERT model, as shown in Figure \ref{fig: cobert}.

\subsection{Task Definition}
% \noindent \textbf{Task Definition} 
We denote a training conversation dataset $\mathcal{D}$ as a list of $N$ conversations in the format of $(X, P, y)$, where $X=\{X_1, X_2, ..., X_{n_X}\}$ denotes the $n_X$ context utterances, $P=\{P_1, P_2, ..., P_{n_P}\}$ denotes the $n_P$ persona sentences of the respondent, and $y$ denotes the response to $X$. The task of response selection can be formulated as learning a function $f(X, P, y)$ that assigns the highest score to the true candidate $y$ and lower scores to negative candidates given $X$ and $P$. During inference, the trained model selects the response candidate with the highest score from a list of candidates.
\subsection{BERT Representation}
% \noindent \textbf{BERT Representation} 
We use BERT \cite{devlin2019bert} as our sentence encoders. Similar to the Bi-encoder \cite{humeau2020polyencoders}, we concatenate context utterances as a single context sentence before passing it into BERT. Since there is no ordering among persona sentences, we concatenate randomly ordered persona sentences\footnote{Reusing the same positional information for all persona sentences \cite{wolf2019transfertransfo} to model position invariance produces worse performance in our preliminary experiments.}. After passing the context, persona and response to BERT encoders, we obtain their vector representations $\mathbf{X} \in \mathbb{R}^{m\times d}$, $\mathbf{P} \in \mathbb{R}^{q\times d}$ and $\mathbf{Y} \in \mathbb{R}^{n\times d}$ from the last layer, respectively, where $d$ denotes the embedding size of BERT, and $m$, $q$ and $n$ denote the sequence lengths of context, persona and response, respectively. Note that different segment ids are used to differentiate speaker and respondent utterances in the context.

\subsection{Hop-$1$ Co-attention}
% \noindent \textbf{Hop-$1$ Co-attention}
Given $\mathbf{X}$ and $\mathbf{Y}$, we learn the first-order matching information using co-attention \cite{lu2016hierarchical}. Specifically, we first compute the word-word affinity matrix $\mathbf{A}_{\mathbf{XY}} \in \mathbb{R}^{m\times n}$:
\begin{equation}
\label{eqn: attention matrix}
    \mathbf{A}_{\mathbf{XY}} = \mathbf{X} \mathbf{Y}^T.
\end{equation}
Then the context-to-response attention $\mathbf{A}_{\mathbf{X2Y}} \in \mathbb{R}^{m\times n}$ and the response-to-context attention $\mathbf{A}_{\mathbf{Y2X}} \in \mathbb{R}^{n\times m}$ can be computed as follows:
\begin{equation}
\label{eqn: hop-1 X}
    \mathbf{A}_{\mathbf{X2Y}} = \textit{softmax}(\mathbf{A}_{\mathbf{XY}}),
\end{equation}
\begin{equation}
\label{eqn: hop-1 Y}
    \mathbf{A}_{\mathbf{Y2X}} = \textit{softmax}(\mathbf{A}_{\mathbf{XY}}^T),
\end{equation}
where $\textit{softmax}$ denotes the softmax function along the second dimension. Finally, we obtain the attended context representation $\mathbf{X}^{'} = \mathbf{A}_{\mathbf{X2Y}} \mathbf{Y} \in \mathbb{R}^{m\times d}$ and response representation $\mathbf{Y}_{\mathbf{X}}^{'} = \mathbf{A}_{\mathbf{Y2X}} \mathbf{X} \in \mathbb{R}^{n\times d}$.

To aggregate the first-order matching information and extract discriminative features, we apply max-pooling to $\mathbf{X}^{'}$ and $\mathbf{Y}_{\mathbf{X}}^{'}$ along the sequence dimension and obtain $\mathbf{X}^{'}_{max} \in \mathbb{R}^{d}$ and $\mathbf{Y}_{\mathbf{X}, max}^{'} \in \mathbb{R}^{d}$.

\subsection{Hop-$2$ Co-attention}
% \noindent \textbf{Hop-$2$ Co-attention}
We propose a hop-$2$ co-attention to learn second-order interactive matching. Different from the attention-over-attention for reading comprehension \cite{cui2017attention}, our method learns bidirectional matching for response selection. Specifically, we apply attention over the attention matrices:
\begin{equation}
\label{eqn: hop-2 X}
    \mathbf{A_X}^{'} = \textit{mean}(\mathbf{A}_{\mathbf{X2Y}}) \mathbf{A}_{\mathbf{Y2X}},
\end{equation}
\begin{equation}
\label{eqn: hop-2 Y}
    \mathbf{A_Y}^{'} = \textit{mean}(\mathbf{A}_{\mathbf{Y2X}}) \mathbf{A}_{\mathbf{X2Y}},
\end{equation}
where $\mathbf{A_X}^{'} \in \mathbb{R}^{1\times m}$ and $\mathbf{A_Y}^{'} \in \mathbb{R}^{1\times n}$ denote the second-order attention over $\mathbf{X}$ and  $\mathbf{Y}$, respectively, and \textit{mean} denotes mean pooling along the first dimension. Then we obtain the attended context representation $\mathbf{X}^{''} = \mathbf{A_X}^{'} \mathbf{X} \in \mathbb{R}^{d}$ and response representation $\mathbf{Y}_{\mathbf{X}}^{''} = \mathbf{A_Y}^{'} \mathbf{Y} \in \mathbb{R}^{d}$. 

We apply the same procedure to match $\mathbf{P}$ and $\mathbf{Y}$, and obtain the first-order matching information $\mathbf{P}^{'}_{max} \in \mathbb{R}^{d}$ and $\mathbf{Y}_{\mathbf{P}, max}^{'} \in \mathbb{R}^{d}$, and the second-order matching information $\mathbf{P}^{''} \in \mathbb{R}^{d}$ and $\mathbf{Y}_{\mathbf{P}}^{''} \in \mathbb{R}^{d}$.

Intuitively, our hop-$1$ co-attention learns attended representations for $\mathbf{X}$ and $\mathbf{Y}$, and our hop-$2$ co-attention learns ``truly" attended representations for $\mathbf{X}$ and $\mathbf{Y}$ where the weights are computed from attentions over attentions.

\subsection{Loss}
% \noindent \textbf{Loss}
We obtain the final persona-aware context representation $\mathbf{X}_f = [\mathbf{X}^{'}_{max}; \mathbf{X}^{''}; \mathbf{P}^{'}_{max}; \mathbf{P}^{''}] \in \mathbb{R}^{4d}$ and the final response representation $\mathbf{Y}_f = [\mathbf{Y}_{\mathbf{X}, max}^{'}; \mathbf{Y}_{\mathbf{X}}^{''}; \mathbf{Y}_{\mathbf{P}, max}^{'}; \mathbf{Y}_{\mathbf{P}}^{''}] \in \mathbb{R}^{4d}$, where $[;]$ denotes concatenation. Then we use dot product to compute the final matching score: 
\begin{equation}
    f(X,P,y) = \textit{dot}(\mathbf{X}_f, \mathbf{Y}_f).
\end{equation}
We optimize our model by minimizing the cross-entropy loss for selecting the true candidate from a list of candidates. Formally, the loss $\Phi$ is computed as follows:
\begin{equation}
\label{eqn: loss}
    \Phi = \sum_{(X,P,y)\thicksim \mathcal{D}}-\frac{e^{f(X,P,y)}}{\sum_{\hat{y} \thicksim {\mathcal{N}(X)} \cup \{y\}} e^{f(X,P,\hat{y})}},
\end{equation}
where $\mathcal{N}(X)$ denotes a set of randomly sampled negative candidates for the context $X$.

\section{Experiments}
\label{sec: experiments}
In this section we present the datasets, baselines, experimental settings, model comparisons and ablation studies.

\subsection{Datasets and Baselines}
\begin{table*}[!t]
\small
\centering
\begin{tabular}{P{2cm}|P{0.68cm}P{0.68cm}P{0.68cm}P{0.68cm}|P{0.68cm}P{0.68cm}P{0.68cm}P{0.68cm}|P{0.68cm}P{0.68cm}P{0.68cm}P{0.68cm}}
% \begin{tabular}{c|c|c|c|c|c|c|c|c|c|c|c|c}
\hline
 & \multicolumn{4}{c|}{\textbf{happy}} & \multicolumn{4}{c|}{\textbf{offmychest}} &
 \multicolumn{4}{c}{\textbf{PEC (happy + offmychest)}}\\
\hline
\textbf{Models} & R@$1$ & R@$10$ & R@$50$ & MRR & R@$1$ & R@$10$ & R@$50$ & MRR &R@$1$ & R@$10$ & R@$50$ & MRR\\
\hline
BoW & 10.2 & 45.6 & 85.2 & 21.8 & 13.9 & 51.6 & 87.1 & 26.2 & 15.4 & 52.9 & 86.7 & 27.4 \\
HLSTM & 15.7 & 53.6 & 91.6 & 28.1 & 17.6 & 55.7 & 91.8 & 30.2 & 22.2 & 63.0 & 94.8 & 35.2 \\
DIM & 31.3 & 67.0 & 95.5 & 43.0 & 40.6 & 72.6 & 96.4 & 51.2 & 39.3 & 74.6 & 97.3 & 50.5 \\
Bi-encoder & 32.4 & 71.3 & 96.5 & 45.1 & 42.4 & 78.4 & 97.6 & 54.5 & 42.3 & 79.2 & 98.1 & 54.4 \\
Poly-encoder & 33.7 & 72.1 & 96.7 & 46.4 & 43.4 & 79.3 & 97.7 & 55.3 & 43.0 & 79.8 & 98.2 & 55.2 \\
\hline
CoBERT (ours) & \textbf{36.2} & \textbf{73.0} & \textbf{96.9} & \textbf{48.4} & \textbf{47.0} & \textbf{79.7} & \textbf{97.8} & \textbf{58.0} & \textbf{45.1} & \textbf{80.5} & \textbf{98.3} & \textbf{56.7} \\
\hline
\end{tabular}
\caption{Test performance (in \%) of CoBERT and all baselines. Values in bold denote best results.}
\label{table: test results}
\end{table*}
We evaluate models on PEC and its two sub-domains, i.e., happy and offmychest. The training, validation and test splits of PEC are combined from the corresponding splits from happy and offmychest. The dataset statistics are shown in Table~\ref{table: datasets}. 

We compare CoBERT with several competitive baselines. 
Note that the BoW, HLSTM \cite{lowe2015ubuntu} and Bi-encoder \cite{humeau2020polyencoders} baselines share the same Tri-encoder architecture, where the final matching score is the dot product between the average of context and persona representations and the response representation. 

\noindent \textbf{BoW}: The context, persona and response encoders compute the averaged word embedding.

\noindent \textbf{HLSTM} \cite{lowe2015ubuntu}: The context encoder has an utterance-level BiLSTM and a context-level BiLSTM. All encoders share the same utterance-level BiLSTM. 

\noindent \textbf{DIM} \cite{gu2019dually}: A state-of-the-art non-pretraiend model for persona-based response selection. DIM adopts finer-grained matching and hierarchical aggregation to learn rich matching representation.

\noindent \textbf{Bi-encoder} \cite{humeau2020polyencoders}: A state-of-the-art BERT-based model for empathetic response selection \cite{rashkin2019towards}.

\noindent \textbf{Poly-encoder} \cite{humeau2020polyencoders}: A state-of-the-art BERT-based model for response selection. Poly-encoder learns latent attention codes for finer-grained matching. Note that we do not consider Cross-encoder \cite{humeau2020polyencoders} as an appropriate baseline because it performs two orders of magnitude slower than Poly-encoder in inference, rendering it intractable for real-time applications.

\subsection{Experimental Settings}
\noindent \textbf{Model Settings} We use fastText \cite{paszke2019pytorch} embeddings of size 300 to initialize  BoW and HLSTM. We follow the released code\footnote{https://github.com/JasonForJoy/DIM} to implement DIM. For all BERT-based models, we use the base version of BERT and share parameters across all three encoders\footnote{A shared BERT encoder obtained better performance than separate encoders in our preliminary experiments.}. We use 128 context codes for Poly-encoder\footnote{More context codes result in memory error in our experiments. According to \cite{humeau2020polyencoders}, more context codes only lead to marginally better results.}. We optimize all BERT-based models using Adam \cite{kingma2014adam} with batch size of 64 and learning rate of 0.00002. The positive to negative candidates ratio during training is set to 1:15. We use a maximum of $n_X=6$ contextual utterances and a maximum of $n_P = 10$ persona sentences for each conversation. We conduct all experiments on NVIDIA V100 32GB GPUs in mixed precision. 

\noindent \textbf{Evaluation Metrics} Following \cite{zhou2018multi, gu2019dually, humeau2020polyencoders}, we evaluate models using Recall@$k$ where each test example has $C$ possible candidates to select from, abbreviated to R@$k$, as well as mean reciprocal rank (MRR). In our experiments, we set $C=100$ and $k=1,10,50$. The candidate set for each test example includes the true response and other $C-1$ randomly sampled responses from the test set.
\begin{table}[!t]
\small
\centering
\begin{tabular}{c|ccc}
\hline
 \diagbox{Train}{Test} & happy & offmychest & PEC\\
\hline
 happy & 36.2 & 41.2 & 40.5\\
 offmychest & 28.8 & 47.0 & 38.4\\
 PEC & 37.0 & 47.5 & 45.1\\
\hline
\end{tabular}
\caption{Transfer test of CoBERT in R@$1$ (in \%).}
\label{table: RQ1}
\end{table}
\subsection{Comparison with Baselines}
We report the test results of response selection in Table \ref{table: test results}. Among the non-pretrained models, DIM outperforms BoW and HLSTM by large margins on all datasets, demonstrating the importance of finer-grained matching and hierarchical aggregation for response selection. 
The simple Bi-encoder performs noticeably better than DIM, suggesting that sentence representation is another critical factor in response selection and that BERT can provide much richer representation than the BiLSTM used in DIM.
Poly-encoder performs best among all baselines because it leverages the strengths of both BERT and attention-based finer-grained matching. 

Our CoBERT consistently outperforms all baselines on all datasets with large margins, including the state-of-the-art Poly-encoder. The performance gain is primarily attributed to our multi-hop co-attention, which learns higher-order bidirectional word-word matching between context and response, whereas Poly-encoder only learns the first-order unidirectional attention from response to context using latent attention codes.
Efficiency-wise, CoBERT has slightly longer inference time (1.50x) but requires much less memory usage (0.62x) than Poly-encoder, as shown in Table \ref{table: ablation study}.

We further investigate the transfer performance of CoBERT in Table \ref{table: RQ1}. In general, in-domain test results are better than out-of-domain test results. 
The transfer performance from happy to offmychest (41.2\%) and vice versa (28.8\%) are comparable to the in-domain performance of DIM (40.6\% on offmychest and 31.3\% on happy), suggesting that our CoBERT can generalize well across empathetic conversations in contrasting sentiments.
\begin{table}[!t]
\small
\centering
\begin{tabular}{ccccc}
\hline
Model & R@$1$ & MRR & InfTime & RAM\\
\hline
\multicolumn{5}{c}{Baselines} \\
\hline
DIM & 40.3 & 51.6 & 10.36x & \textbf{0.79x}\\
Bi-encoder & 42.6 & 55.2 & 1.00x & 1.00x\\
Poly-encoder & 43.3 & 55.7 & 1.33x & 1.84x\\
% BERT+DAM & 45.4 & 57.0 & 10.77x \\
\hline
\multicolumn{5}{c}{BERT-adapted Models} \\
\hline
BERT+MemNet & 42.3 & 53.8 & \textbf{0.87x} & 0.89x\\
BERT+DAM & 45.0 & 56.9 & 14.26x & 1.57x\\
BERT+DIM & 46.1 & 57.7 & 18.36x & 1.78x\\
\hline
\multicolumn{5}{c}{Ablations} \\
\hline
CoBERT (ours) & \textbf{46.2} & \textbf{57.9} & 2.00x & 1.14x\\
- hop-$1$ & 44.0 & 56.2 & 1.65x & 1.11x\\
- hop-$2$ & 45.5 & 57.1 & 1.76x & 1.11x\\
+ hop-$3$ & 46.0 & 57.6 & 2.70x & 1.13x\\
- max + mean & 44.1 & 56.3 & 2.12x & 1.13x\\
+ mean & 46.1 & 57.8 & 2.71x & 1.15x\\
\hline
\end{tabular}
\caption{Validation performance (in \%), inference time (InfTime) and memory usage (RAM) for baselines, BERT-adapted models and ablation studies on PEC. InfTime and RAM are relative to the Bi-encoder.}
\label{table: ablation study}
\end{table}
\subsection{Comparison with BERT-adapted Models}
To perform a more comprehensive evaluation of CoBERT, we further compare CoBERT with several competitive BERT-adapted models where the sentence encoders are replaced by BERT. We report the results in the middle section of Table \ref{table: ablation study}.

\noindent \textbf{BERT + MemNet} \cite{zhang2018personalizing}: MemNet incorporates persona into context using a Memory Network \cite{sukhbaatar2015end} with residual connections. The BERT+MemNet model performs slightly worse than Bi-encoder and much worse than our CoBERT, although it achieves slightly faster inference than Bi-encoder.

\noindent \textbf{BERT+DAM} \cite{zhou2018multi}: DAM aggregates multi-granularity matching using convolutional layers. The BERT+DAM model performs significantly better than Bi-encoder in R@$1$, demonstrating the usefulness of learning n-gram matching over the word-word matching matrices. Nevertheless, CoBERT performs noticeably better and has faster inference (7.13x) than BERT+DAM.

\noindent \textbf{BERT+DIM} \cite{gu2019dually}: 
The BERT+DIM model combines the benefits from both the strong sentence representation of BERT and the rich finer-grained matching of DIM. However, BERT+DIM performs slightly worse than CoBERT, suggesting that the more complex matching and aggregation methods in DIM do not lead to performance improvement over our multi-hop co-attention. In addition, our CoBERT is substantially faster (9.18x) than BERT+DIM in inference, thus more practical in real-world applications.

\subsection{Ablation Study}
\label{sec: ablation study}
We conduct ablation studies for CoBERT, as reported in the bottom section of Table \ref{table: ablation study}. 

Removing either hop-1 or hop-2 co-attention results in noticeably worse performance, albeit slightly faster inference. Removing hop-1 leads to larger performance drop than removing hop-2, suggesting that the first-order matching information seems more important than the second-order matching information for response selection. An additional hop-3 co-attention results in slightly worse performance, suggesting that our two-hop co-attention is the sweet spot for model complexity. 

Replacing the max pooling in the hop-1 co-attention by mean pooling leads to much worse performance. In addition, concatenating the results from both max and mean pooling slightly degrades performance, as well as inference speed, suggesting that max pooling may be essential for extracting discriminative matching information.

\section{Discussion}
\label{sec: discussion}
\begin{figure}
    \centering
    \includegraphics[width=\linewidth]{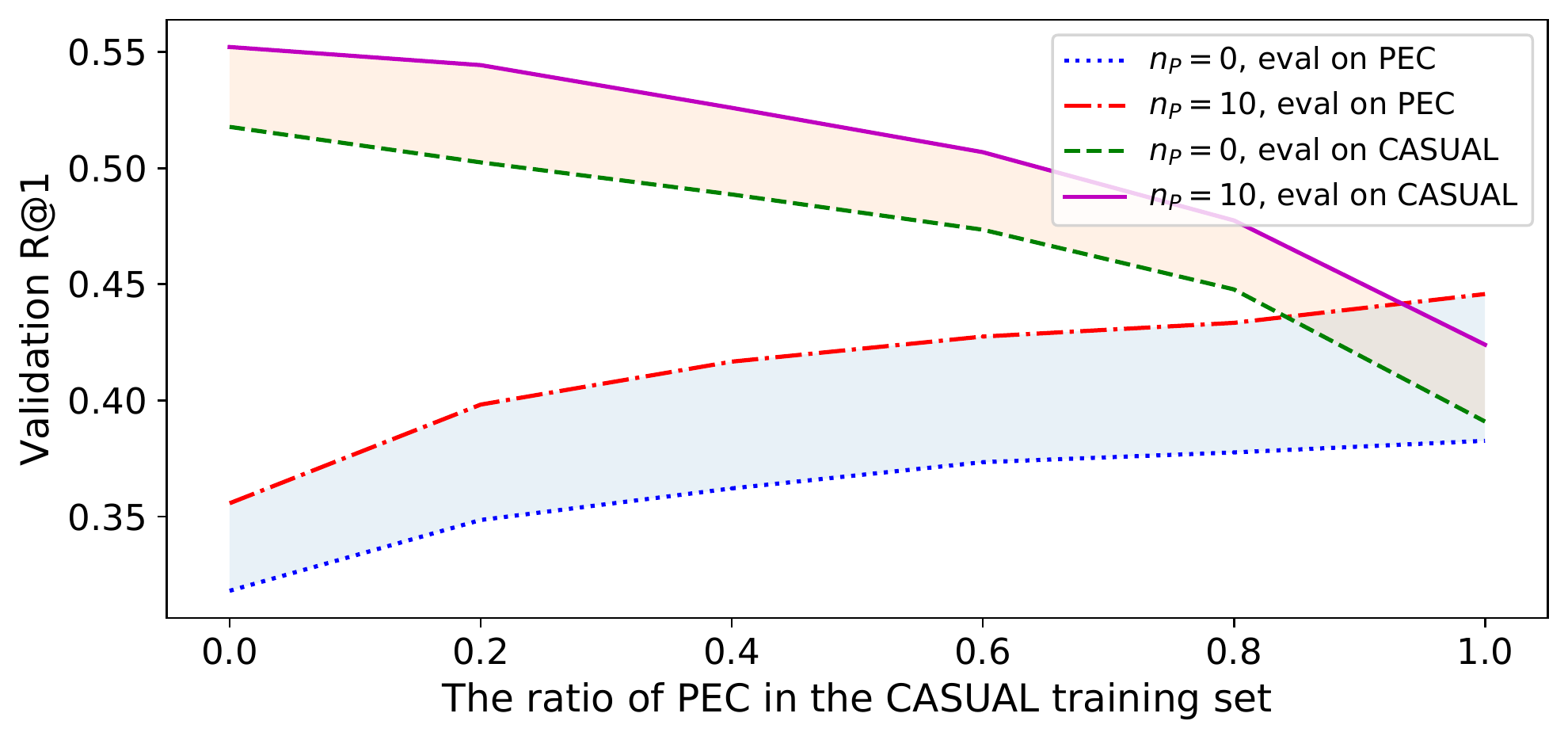}
    \caption{Validation R@$1$ (in \%) against different ratios of PEC in the CASUAL training set.}
    \label{fig: RQ2}
\end{figure}
\subsection{Empathetic vs. Non-empathetic}
% \noindent\textbf{Empathetic vs. Non-empathetic}
We investigate whether persona improves empathetic responding more when CoBERT is trained on empathetic conversations than non-empathetic ones. 
First, we introduce a non-empathetic conversation dataset as the control group, denoted as CASUAL, which is the same as the control group in Section \ref{sec: PEC dataset} but much larger in size. The CASUAL dataset is collected and processed in the same way as PEC but has significantly lower empathy than PEC (see Table \ref{table: dataset annotations}). The sizes of training, validation, and testing splits of CASUAL are 150K, 20K, and 20K, respectively. 
Then, we replace a random subset of training examples from CASUAL by the same number of random training examples from PEC. We then compare the persona improvement, i.e., R@$1$ ($n_P=10$) $-$ R@$1$ ($n_P=0$), on the PEC validation set and the CASUAL validation set for different replacement ratios. 

The results are illustrated in Figure \ref{fig: RQ2}. It is unsurprising that for both cases, i.e., $n_P=0$ and $n_P=10$, the validation R@$1$ on PEC increases, and the validation R@$1$ on CASUAL decreases as the ratio of PEC in the training dataset increases. We also observe that persona consistently improves performance on both validation sets for all ratios.

By investigating the widths of the two shaded regions in Figure \ref{fig: RQ2}, we find that the persona improvement on casual responding remains almost constant as more CASUAL training examples are used (3.31\% when trained on all 150K PEC conversations vs. 3.44\% when trained on all 150K CASUAL conversations). However, the persona improvement on empathetic responding consistently increases as more PEC training examples are used (3.77\% when trained on all 150K CASUAL conversations versus 6.32\% when trained on all 150K PEC conversations), showing that persona improves empathetic responding significantly more when CoBERT is trained on empathetic conversations than non-empathetic ones ($p<0.001$, one-tailed $t$-test). 

This result reveals an empirical link between persona and empathy in human conversations and may suggest that persona has a greater impact on empathetic conversations than non-empathetic ones. The result also shows that CoBERT can learn this link during training and use it to perform better empathetic responding during testing. One possible psychological root of this link is that persona is highly correlated to personality \cite{leary2011personality}, which in turn influences empathy and empathetic responding \cite{costa2014associations}. 
A more detailed analysis of this empirical link is left for future work.

\subsection{Number of Persona Sentences}
% \noindent\textbf{Number of Persona Sentences}
We analyze the persona improvement with respect to different numbers of persona sentences $n_P$, as shown in Table \ref{table: RQ4}\footnote{Using $n_P=30$ results in memory error.}. It is clear that model performance, inference time, and memory usage all increase when more persona sentences are incorporated. Note that memory usage grows quadratically with $n_P$ due to the self-attention operations in BERT. 
We chose $n_P=10$ in our experiments because it achieves competitive performance at a reasonable cost of efficiency.
\begin{table}[!t]
\small
\centering
\begin{tabular}{P{0.9cm}P{0.6cm}P{0.6cm}P{0.6cm}P{0.6cm}P{0.6cm}P{0.6cm}}
% \begin{tabular}{c|c|c|c|c|c|c}
\hline
$n_P$ & 0 & 1 & 2 & 5 & 10 & 20\\
\hline
R@$1$  & 40.4 & 42.0 & 42.8 & 45.1 & 46.2 & \textbf{47.1}\\
InfTime  & \textbf{1.00x} & 1.34x & 1.38x & 1.55x & 1.90x & 2.96x\\
RAM  & \textbf{1.00x} & 1.05x & 1.06x & 1.19x & 1.51x & 2.29x\\
\hline
\end{tabular}
\caption{Validation R@$1$ (in \%), inference time (InfTime) and memory usage (RAM) on PEC against different number of persona sentences $n_P$.}
\label{table: RQ4}
\end{table}
\begin{table}[!t]
\small
\centering
% \begin{tabular}{P{0.9cm}P{0.6cm}P{0.6cm}P{0.6cm}P{0.6cm}P{0.6cm}P{0.6cm}}
\begin{tabular}{cccc}
\hline
$n_P$ & seen (57.9\%) & unseen (42.1\%) & all (100\%) \\
\hline
0 & 40.3 & 38.5 & 39.6\\
10 & 46.5 & 43.2 & 45.1\\
\hline
\end{tabular}
\caption{Test R@$1$ (in \%) on PEC against examples with seen or unseen personas. $n_P$ denotes the number of persona sentences.}
\label{table: new speakers}
\end{table}
\subsection{Performance on New Personas}
% \noindent\textbf{New Personas}
We analyze the CoBERT performance on examples with new personas. In PEC test set, 42.1\% examples are from new speakers. The performance of CoBERT on test examples with seen and unseen (new) speakers is shown in Table \ref{table: new speakers}. The results show that 1) CoBERT performs reasonably well on examples with unseen personas, suggesting that CoBERT can generalize well to unseen personas and retrieve the right response for new speakers accurately; 2) CoBERT performs worse on examples with unseen personas than seen personas; 3) leveraging personas during model training and testing improves CoBERT on examples with either seen or unseen personas; and 4) the persona improvement is more noticeable for examples with seen personas than unseen personas.
\begin{table}[!t]
\small
\centering
\begin{tabular}{p{2cm}|p{4.8cm}}
% \begin{tabular}{c|c}
\hline
\multicolumn{2}{l}{\textbf{Context}: I'm on a diet and lost ten pounds this month!}\\
\hline
\textbf{Persona} & \textbf{Model Response}\\
\hline
% I'm on a diet and lost ten pounds this month! & \\
I am a college graduate. & Congrats! I'm trying to lose weight as well.\\
\hline
I work in a gym. & Good job! Doing some exercises will help you stay fit!\\
\hline
I am a doctor. & Congrats! Don't forget to take adequate nutrition though.\\
\hline
\end{tabular}
\caption{Case study.}
\label{table: case study}
\end{table}
\subsection{Case Study}
% \noindent\textbf{Case Study} 
We conduct a case study on how persona affects empathetic responding, as shown in Table \ref{table: case study}. The model responses are selected by CoBERT from 1K candidates. It is clear that given the same context, different personas lead to different persona-based empathetic responses. For example, when the persona is ``I am a doctor.", the model response expresses both praises and caring about the speaker's health.

\section{Conclusion}
We present a new task and a large-scale multi-domain dataset, PEC, towards persona-based empathetic conversations. We then propose CoBERT, an effective and efficient model that obtains substantially better performance than competitive baselines on PEC, including the state-of-the-art Poly-encoder and several BERT-adapted models. CoBERT is free from hyper-parameter tuning and universally applicable to the task of response selection in any domain. Finally, we present the first empirical study on the impact of persona on empathetic responding. 
The results reveal an empirical link between persona and empathy in human conversations and may suggest that persona has a greater impact on empathetic conversations than non-empathetic ones. 

\section*{Acknowledgments}
This research is supported, in part, by Alibaba Group through Alibaba Innovative Research (AIR) Program and Alibaba-NTU Singapore Joint Research Institute (JRI) (Alibaba-NTU-AIR2019B1), Nanyang Technological University, Singapore. This research is also supported, in part, by the National Research Foundation, Prime Minister's Office, Singapore under its AI Singapore Programme (AISG Award No: AISG-GC-2019-003) and under its NRF Investigatorship Programme (NRFI Award No. NRF-NRFI05-2019-0002). Any opinions, findings and conclusions or recommendations expressed in this material are those of the authors and do not reflect the views of National Research Foundation, Singapore. This research is also supported, in part, by the Singapore Ministry of Health under its National Innovation Challenge on Active and Confident Ageing (NIC Project No. MOH/NIC/COG04/2017 and MOH/NIC/HAIG03/2017).

\bibliography{emnlp2020}

\begin{thebibliography}{47}
\expandafter\ifx\csname natexlab\endcsname\relax\def\natexlab#1{#1}\fi

\bibitem[{Brave et~al.(2005)Brave, Nass, and Hutchinson}]{brave2005computers}
Scott Brave, Clifford Nass, and Kevin Hutchinson. 2005.
\newblock Computers that care: investigating the effects of orientation of
  emotion exhibited by an embodied computer agent.
\newblock \emph{International Journal of Human-Computer Studies},
  62(2):161--178.

\bibitem[{Chan et~al.(2019)Chan, Li, Yang, Chen, Hu, Zhao, and
  Yan}]{chan2019modeling}
Zhangming Chan, Juntao Li, Xiaopeng Yang, Xiuying Chen, Wenpeng Hu, Dongyan
  Zhao, and Rui Yan. 2019.
\newblock Modeling personalization in continuous space for response generation
  via augmented wasserstein autoencoders.
\newblock In \emph{EMNLP-IJCNLP}, pages 1931--1940.

\bibitem[{Chen and Wang(2019)}]{chen2019sequential}
Qian Chen and Wen Wang. 2019.
\newblock Sequential attention-based network for noetic end-to-end response
  selection.
\newblock \emph{arXiv preprint arXiv:1901.02609}.

\bibitem[{Costa et~al.(2014)Costa, Alves, Neto, Marvao, Portela, and
  Costa}]{costa2014associations}
Patricio Costa, Raquel Alves, Isabel Neto, Pedro Marvao, Miguel Portela, and
  Manuel~Joao Costa. 2014.
\newblock Associations between medical student empathy and personality: a
  multi-institutional study.
\newblock \emph{PloS one}, 9(3).

\bibitem[{Cui et~al.(2017)Cui, Chen, Wei, Wang, Liu, and Hu}]{cui2017attention}
Yiming Cui, Zhipeng Chen, Si~Wei, Shijin Wang, Ting Liu, and Guoping Hu. 2017.
\newblock Attention-over-attention neural networks for reading comprehension.
\newblock In \emph{ACL}, pages 593--602.

\bibitem[{Devlin et~al.(2019)Devlin, Chang, Lee, and
  Toutanova}]{devlin2019bert}
Jacob Devlin, Ming-Wei Chang, Kenton Lee, and Kristina Toutanova. 2019.
\newblock Bert: Pre-training of deep bidirectional transformers for language
  understanding.
\newblock In \emph{NAACL}, pages 4171--4186.

\bibitem[{Feng et~al.(2019)Feng, Tao, Wu, Feng, Zhao, and
  Yan}]{feng2019learning}
Jiazhan Feng, Chongyang Tao, Wei Wu, Yansong Feng, Dongyan Zhao, and Rui Yan.
  2019.
\newblock Learning a matching model with co-teaching for multi-turn response
  selection in retrieval-based dialogue systems.
\newblock In \emph{ACL}, pages 3805--3815.

\bibitem[{Fitzpatrick et~al.(2017)Fitzpatrick, Darcy, and
  Vierhile}]{fitzpatrick2017delivering}
Kathleen~Kara Fitzpatrick, Alison Darcy, and Molly Vierhile. 2017.
\newblock Delivering cognitive behavior therapy to young adults with symptoms
  of depression and anxiety using a fully automated conversational agent
  (woebot): a randomized controlled trial.
\newblock \emph{JMIR Mental Health}, 4(2):e19.

\bibitem[{Fleiss(1971)}]{fleiss1971measuring}
Joseph~L Fleiss. 1971.
\newblock Measuring nominal scale agreement among many raters.
\newblock \emph{Psychological Bulletin}, 76(5):378.

\bibitem[{Gu et~al.(2019)Gu, Ling, Zhu, and Liu}]{gu2019dually}
Jia-Chen Gu, Zhen-Hua Ling, Xiaodan Zhu, and Quan Liu. 2019.
\newblock Dually interactive matching network for personalized response
  selection in retrieval-based chatbots.
\newblock In \emph{EMNLP-IJCNLP}, pages 1845--1854.

\bibitem[{Humeau et~al.(2020)Humeau, Shuster, Lachaux, and
  Weston}]{humeau2020polyencoders}
Samuel Humeau, Kurt Shuster, Marie-Anne Lachaux, and Jason Weston. 2020.
\newblock Poly-encoders: Architectures and pre-training strategies for fast and
  accurate multi-sentence scoring.
\newblock In \emph{ICLR}.

\bibitem[{Jung(2016)}]{jung2016psychological}
Carl Jung. 2016.
\newblock \emph{Psychological types}.
\newblock Taylor \& Francis.

\bibitem[{Kingma and Ba(2014)}]{kingma2014adam}
Diederik~P Kingma and Jimmy Ba. 2014.
\newblock Adam: A method for stochastic optimization.
\newblock \emph{arXiv preprint arXiv:1412.6980}.

\bibitem[{Klein(1998)}]{klein1998computer}
Jonathan~Tarter Klein. 1998.
\newblock \emph{Computer response to user frustration}.
\newblock Ph.D. thesis, Massachusetts Institute of Technology.

\bibitem[{Leary and Allen(2011)}]{leary2011personality}
Mark~R Leary and Ashley~Batts Allen. 2011.
\newblock Personality and persona: Personality processes in self-presentation.
\newblock \emph{Journal of Personality}, 79(6):1191--1218.

\bibitem[{Li et~al.(2016)Li, Galley, Brockett, Spithourakis, Gao, and
  Dolan}]{li2016persona}
Jiwei Li, Michel Galley, Chris Brockett, Georgios Spithourakis, Jianfeng Gao,
  and Bill Dolan. 2016.
\newblock A persona-based neural conversation model.
\newblock In \emph{ACL}, pages 994--1003.

\bibitem[{Li et~al.(2019)Li, Chen, Ren, Chen, Tu, and Ma}]{li2019empgan}
Qintong Li, Hongshen Chen, Zhaochun Ren, Zhumin Chen, Zhaopeng Tu, and Jun Ma.
  2019.
\newblock Emp{GAN}: Multi-resolution interactive empathetic dialogue
  generation.
\newblock \emph{arXiv preprint arXiv:1911.08698}.

\bibitem[{Lin et~al.(2019{\natexlab{a}})Lin, Madotto, Shin, Xu, and
  Fung}]{lin2019moel}
Zhaojiang Lin, Andrea Madotto, Jamin Shin, Peng Xu, and Pascale Fung.
  2019{\natexlab{a}}.
\newblock Moel: Mixture of empathetic listeners.
\newblock In \emph{EMNLP-IJCNLP}, pages 121--132.

\bibitem[{Lin et~al.(2019{\natexlab{b}})Lin, Xu, Winata, Liu, and
  Fung}]{lin2019caire}
Zhaojiang Lin, Peng Xu, Genta~Indra Winata, Zihan Liu, and Pascale Fung.
  2019{\natexlab{b}}.
\newblock Caire: An end-to-end empathetic chatbot.
\newblock \emph{arXiv preprint arXiv:1907.12108}.

\bibitem[{Liu and Picard(2005)}]{liu2005embedded}
K~Liu and Rosalind~W Picard. 2005.
\newblock Embedded empathy in continuous, interactive health assessment.
\newblock In \emph{CHI Workshop on HCI Challenges in Health Assessment},
  volume~1, page~3.

\bibitem[{Lowe et~al.(2015)Lowe, Pow, Serban, and Pineau}]{lowe2015ubuntu}
Ryan Lowe, Nissan Pow, Iulian~Vlad Serban, and Joelle Pineau. 2015.
\newblock The ubuntu dialogue corpus: A large dataset for research in
  unstructured multi-turn dialogue systems.
\newblock In \emph{SIGDIAL}, pages 285--294.

\bibitem[{Lu et~al.(2016)Lu, Yang, Batra, and Parikh}]{lu2016hierarchical}
Jiasen Lu, Jianwei Yang, Dhruv Batra, and Devi Parikh. 2016.
\newblock Hierarchical question-image co-attention for visual question
  answering.
\newblock In \emph{NIPS}, pages 289--297.

\bibitem[{Madotto et~al.(2019)Madotto, Lin, Wu, and
  Fung}]{madotto2019personalizing}
Andrea Madotto, Zhaojiang Lin, Chien-Sheng Wu, and Pascale Fung. 2019.
\newblock Personalizing dialogue agents via meta-learning.
\newblock In \emph{ACL}, pages 5454--5459.

\bibitem[{Mazare et~al.(2018)Mazare, Humeau, Raison, and
  Bordes}]{mazare2018training}
Pierre-Emmanuel Mazare, Samuel Humeau, Martin Raison, and Antoine Bordes. 2018.
\newblock Training millions of personalized dialogue agents.
\newblock In \emph{EMNLP}, pages 2775--2779.

\bibitem[{Morris et~al.(2018)Morris, Kouddous, Kshirsagar, and
  Schueller}]{morris2018towards}
Robert~R Morris, Kareem Kouddous, Rohan Kshirsagar, and Stephen~M Schueller.
  2018.
\newblock Towards an artificially empathic conversational agent for mental
  health applications: system design and user perceptions.
\newblock \emph{Journal of Medical Internet Research}, 20(6):e10148.

\bibitem[{Paszke et~al.(2019)Paszke, Gross, Massa, Lerer, Bradbury, Chanan,
  Killeen, Lin, Gimelshein, Antiga et~al.}]{paszke2019pytorch}
Adam Paszke, Sam Gross, Francisco Massa, Adam Lerer, James Bradbury, Gregory
  Chanan, Trevor Killeen, Zeming Lin, Natalia Gimelshein, Luca Antiga, et~al.
  2019.
\newblock Pytorch: An imperative style, high-performance deep learning library.
\newblock In \emph{NIPS}, pages 8024--8035.

\bibitem[{Rashkin et~al.(2019)Rashkin, Smith, Li, and
  Boureau}]{rashkin2019towards}
Hannah Rashkin, Eric~Michael Smith, Margaret Li, and Y-Lan Boureau. 2019.
\newblock Towards empathetic open-domain conversation models: A new benchmark
  and dataset.
\newblock In \emph{ACL}, pages 5370--5381.

\bibitem[{Richendoller and Weaver~III(1994)}]{richendoller1994exploring}
Nadine~R Richendoller and James~B Weaver~III. 1994.
\newblock Exploring the links between personality and empathic response style.
\newblock \emph{Personality and Individual Differences}, 17(3):303--311.

\bibitem[{Rogers et~al.(2007)Rogers, Dziobek, Hassenstab, Wolf, and
  Convit}]{rogers2007cares}
Kimberley Rogers, Isabel Dziobek, Jason Hassenstab, Oliver~T Wolf, and Antonio
  Convit. 2007.
\newblock Who cares? revisiting empathy in asperger syndrome.
\newblock \emph{Journal of Autism and Developmental Disorders}, 37(4):709--715.

\bibitem[{Roller et~al.(2020)Roller, Dinan, Goyal, Ju, Williamson, Liu, Xu,
  Ott, Shuster, Smith et~al.}]{roller2020recipes}
Stephen Roller, Emily Dinan, Naman Goyal, Da~Ju, Mary Williamson, Yinhan Liu,
  Jing Xu, Myle Ott, Kurt Shuster, Eric~M Smith, et~al. 2020.
\newblock Recipes for building an open-domain chatbot.
\newblock \emph{arXiv preprint arXiv:2004.13637}.

\bibitem[{Shi and Yu(2018)}]{shi2018sentiment}
Weiyan Shi and Zhou Yu. 2018.
\newblock Sentiment adaptive end-to-end dialog systems.
\newblock In \emph{ACL}, pages 1509--1519.

\bibitem[{Shin et~al.(2019)Shin, Xu, Madotto, and Fung}]{shin2019happybot}
Jamin Shin, Peng Xu, Andrea Madotto, and Pascale Fung. 2019.
\newblock Happybot: Generating empathetic dialogue responses by improving user
  experience look-ahead.
\newblock \emph{arXiv preprint arXiv:1906.08487}.

\bibitem[{Siddique et~al.(2017)Siddique, Kampman, Yang, Dey, and
  Fung}]{siddique2017zara}
Farhad~Bin Siddique, Onno Kampman, Yang Yang, Anik Dey, and Pascale Fung. 2017.
\newblock Zara returns: Improved personality induction and adaptation by an
  empathetic virtual agent.
\newblock In \emph{ACL}, pages 121--126.

\bibitem[{Sukhbaatar et~al.(2015)Sukhbaatar, Weston, Fergus
  et~al.}]{sukhbaatar2015end}
Sainbayar Sukhbaatar, Jason Weston, Rob Fergus, et~al. 2015.
\newblock End-to-end memory networks.
\newblock In \emph{NIPS}, pages 2440--2448.

\bibitem[{Vaswani et~al.(2017)Vaswani, Shazeer, Parmar, Uszkoreit, Jones,
  Gomez, Kaiser, and Polosukhin}]{vaswani2017attention}
Ashish Vaswani, Noam Shazeer, Niki Parmar, Jakob Uszkoreit, Llion Jones,
  Aidan~N Gomez, {\L}ukasz Kaiser, and Illia Polosukhin. 2017.
\newblock Attention is all you need.
\newblock In \emph{NIPS}, pages 5998--6008.

\bibitem[{Vinyals and Le(2015)}]{vinyals2015neural}
Oriol Vinyals and Quoc Le. 2015.
\newblock A neural conversational model.
\newblock \emph{arXiv preprint arXiv:1506.05869}.

\bibitem[{Wolf et~al.(2019)Wolf, Sanh, Chaumond, and
  Delangue}]{wolf2019transfertransfo}
Thomas Wolf, Victor Sanh, Julien Chaumond, and Clement Delangue. 2019.
\newblock Transfertransfo: A transfer learning approach for neural network
  based conversational agents.
\newblock \emph{arXiv preprint arXiv:1901.08149}.

\bibitem[{Wright and McCarthy(2008)}]{wright2008empathy}
Peter Wright and John McCarthy. 2008.
\newblock Empathy and experience in hci.
\newblock In \emph{Proceedings of the SIGCHI Conference on Human Factors in
  Computing Systems}, pages 637--646.

\bibitem[{Wu et~al.(2017)Wu, Wu, Xing, Zhou, and Li}]{wu2017sequential}
Yu~Wu, Wei Wu, Chen Xing, Ming Zhou, and Zhoujun Li. 2017.
\newblock Sequential matching network: A new architecture for multi-turn
  response selection in retrieval-based chatbots.
\newblock In \emph{ACL}, pages 496--505.

\bibitem[{Yuan et~al.(2019)Yuan, Zhou, Li, Lv, Zhu, Han, and
  Hu}]{yuan2019multi}
Chunyuan Yuan, Wei Zhou, Mingming Li, Shangwen Lv, Fuqing Zhu, Jizhong Han, and
  Songlin Hu. 2019.
\newblock Multi-hop selector network for multi-turn response selection in
  retrieval-based chatbots.
\newblock In \emph{EMNLP-IJCNLP}, pages 111--120.

\bibitem[{Zandie and Mahoor(2020)}]{zandie2020emptransfo}
Rohola Zandie and Mohammad~H Mahoor. 2020.
\newblock Emptransfo: A multi-head transformer architecture for creating
  empathetic dialog systems.
\newblock \emph{arXiv preprint arXiv:2003.02958}.

\bibitem[{Zhang et~al.(2018{\natexlab{a}})Zhang, Dinan, Urbanek, Szlam, Kiela,
  and Weston}]{zhang2018personalizing}
Saizheng Zhang, Emily Dinan, Jack Urbanek, Arthur Szlam, Douwe Kiela, and Jason
  Weston. 2018{\natexlab{a}}.
\newblock Personalizing dialogue agents: I have a dog, do you have pets too?
\newblock In \emph{ACL}, pages 2204--2213.

\bibitem[{Zhang et~al.(2018{\natexlab{b}})Zhang, Li, Zhu, Zhao, and
  Liu}]{zhang2018modeling}
Zhuosheng Zhang, Jiangtong Li, Pengfei Zhu, Hai Zhao, and Gongshen Liu.
  2018{\natexlab{b}}.
\newblock Modeling multi-turn conversation with deep utterance aggregation.
\newblock In \emph{COLING}, pages 3740--3752.

\bibitem[{Zheng et~al.(2019)Zheng, Zhang, Mao, and Huang}]{zheng2019pre}
Yinhe Zheng, Rongsheng Zhang, Xiaoxi Mao, and Minlie Huang. 2019.
\newblock A pre-training based personalized dialogue generation model with
  persona-sparse data.
\newblock \emph{arXiv preprint arXiv:1911.04700}.

\bibitem[{Zhou et~al.(2018{\natexlab{a}})Zhou, Gao, Li, and
  Shum}]{zhou2018design}
Li~Zhou, Jianfeng Gao, Di~Li, and Heung-Yeung Shum. 2018{\natexlab{a}}.
\newblock The design and implementation of xiaoice, an empathetic social
  chatbot.
\newblock \emph{Computational Linguistics}, 0:1--62.

\bibitem[{Zhou et~al.(2016)Zhou, Dong, Wu, Zhao, Yu, Tian, Liu, and
  Yan}]{zhou2016multi}
Xiangyang Zhou, Daxiang Dong, Hua Wu, Shiqi Zhao, Dianhai Yu, Hao Tian, Xuan
  Liu, and Rui Yan. 2016.
\newblock Multi-view response selection for human-computer conversation.
\newblock In \emph{EMNLP}, pages 372--381.

\bibitem[{Zhou et~al.(2018{\natexlab{b}})Zhou, Li, Dong, Liu, Chen, Zhao, Yu,
  and Wu}]{zhou2018multi}
Xiangyang Zhou, Lu~Li, Daxiang Dong, Yi~Liu, Ying Chen, Wayne~Xin Zhao, Dianhai
  Yu, and Hua Wu. 2018{\natexlab{b}}.
\newblock Multi-turn response selection for chatbots with deep attention
  matching network.
\newblock In \emph{ACL}, pages 1118--1127.

\end{thebibliography}
\bibliographystyle{acl_natbib}

\end{document}